\documentclass{article}

\usepackage[preprint,nonatbib]{neurips_2020}

\usepackage[utf8]{inputenc} 
\usepackage[T1]{fontenc}    
\usepackage{hyperref}       
\usepackage{url}            
\usepackage{booktabs}       
\usepackage{amsfonts}       
\usepackage{nicefrac}       
\usepackage{microtype}      
\usepackage{graphicx}
\usepackage{wrapfig}
\usepackage[ruled,vlined]{algorithm2e}
\usepackage{newfloat}
\DeclareFloatingEnvironment[fileext=lop]{Algorithm}

\title{An Efficient Shared-memory Parallel Sinkhorn-Knopp Algorithm to Compute the Word Mover's Distance}

\author{%
  Jesmin Jahan Tithi \\
  Parallel Computing Labs\\
  Intel, Santa Clara, CA, USA\\
  \texttt{jesmin.jahan.tithi@intel.com} \\
  \And
  Fabrizio Petrini \\
  Parallel Computing Labs\\
  Intel, Santa Clara, CA, USA\\
  \texttt{fabrizio.petrini@intel.com}
}

\begin{document}

\maketitle
\begin{abstract}
The Word Mover's Distance (WMD) is a metric that measures the semantic dissimilarity between two text documents by computing the cost of optimally moving all words of a source/query document to the most similar words of a target document. Computing WMD between two documents is costly because it requires solving an optimization problem that costs \(O(V^3log(V))\) where \(V\) is the number of unique words in the document. Fortunately, WMD can be framed as an Earth Mover's Distance (EMD) for which the algorithmic complexity can be reduced to \(O(V^2)\) by adding an entropy penalty to the optimization problem and solve it using the Sinkhorn-Knopp Algorithm. Additionally, the computation can be made highly parallel by computing WMD of a single query document against multiple target documents at once (e.g., finding whether a given tweet is similar to any other tweets of a given day). In this paper, we present a shared-memory parallel Sinkhorn-Knopp Algorithm to compute the WMD of one document against many other documents by adopting the \(O(V^2)\) EMD algorithm. We then algorithmically transform the original \(O(V^2)\) dense compute-heavy kernel to a sparse compute kernel and obtain \(67\times\) speedup using \(96\) cores on the state-of-the-art of Intel\textregistered{} 4-sockets Cascade Lake machine w.r.t. its sequential run. Our parallel algorithm is over \(700\times\) faster than the naive parallel python code that internally uses parallel matrix library calls. 
\end{abstract}

\section{Introduction}
The Word Mover's Distance (WMD) \cite{WMD} measures dissimilarity between two text documents as the minimum distance that the embedded words of one document need to ``travel" to reach the embedded words of another document. In other words, WMD captures the extent of changes needed in the words of one document to become like to words of another document. To represent words, WMD uses the word-embedding vector (e.g., word2Vec\cite{word2vec}) learned via neural networks that encode a semantic meaning of the words from their local co-occurrences in sentences. Using word2vec, it is expected that the distance between words Japan and sushi (i.e., \(embeddings(Japan)\) - \(embeddings(sushi)\)) will be similar to the distance between words Bangladesh and biriyani (i.e., \(embeddings(Bangladesh)\)-\(embeddings(biriyani) \)) since their semantic relationships are similar\cite{word2vec}. 

Computing WMD between two documents requires solving an optimization problem that costs \(O(V^3log(V))\) where \(V\) is the number of unique words in the documents. Article \cite{Cuturi:2013:SDL:2999792.2999868} presents an approximation to the Earth Mover's Distance (EMD) reducing the cost per query to \(O(V^2)\). It uses an entropy penalty to the optimization problem that encourages the solution to lie `close' to a transportation plan that sends equal mass from each point in document \(1\) to each point in document \(2\). It then uses the Sinkhorn-Knopp matrix scaling algorithm \cite{Knight:2008:SAC:1404637.1404647} to solve the optimization problem. 
A similar idea can be adapted to compute WMD efficiently. Furthermore, the computation can be made highly parallel by computing WMD of single query/source document against multiple target documents at once. A practical use case of this is to find whether a tweet is similar to any other tweets in a given day. 

The WMD has recently being used in many types of text analysis applications such as: Agglomerative short text clustering \cite{10.1007/978-3-030-14799-0_11}, evaluations of generated texts, for document retrival\cite{brokos2016using}, for machine translation \cite{zhang2016building}, to create document embedding \cite{wu2018word}, automatic essay evaluations \cite{tashu2018pair}. Using WMD as a distance metric provides unprecedented low k-nearest neighbor document classification error rate compared to other standard metrics such as bag-of-words (BOW), term frequency-inverse document frequency (TFIDF), Latent Semantic Indexing (LSI), and Latent Dirichlet Allocation (LDA) \cite{Li:2019:CES:3308558.3313397}. Hence, there is a significant interest in making computation of WMD faster. We take a step forward in this direction by designing an efficient parallel WMD algorithm that requires \(O(V^2/p)\) time per query where \(V\) is the number of words in the vocabulary and \(p\) is the number of threads used.
\paragraph{\bf{Contributions:}}
In this paper, we present a new sparse algorithm to compute Word Movers Distance with better asymptotic theoretical bounds. The presented algorithm is simple enough to implement and also practically efficient. The parallel version scales well on multicores and runs two orders of magnitude faster compared to a python implementation on state-of-the-art CPUs.  

We make the following contributions:
\begin{itemize}
\item {\bf Parallelization:} We present a shared-memory parallel algorithm to compute the WMD of one document against many documents at once by combining a fast EMD presented in \cite{Cuturi:2013:SDL:2999792.2999868} and the Sinkhorn-Knopp matrix-scaling algorithm. 
\item {\bf Algorithmic Innovation:} The resulting kernels from the combination of EMD and Sinkhorn-Knopp are dense, and we algorithmically transform that to a sparse one. We propose a new kernel called SDDMM\_SpMM that fuses the Sparse x Dense x Dense matrix multiplication (SDDMM) and the Dense x Sparse matrix multiplication (SpMM) kernels.
\item {\bf Theoretical Analysis:} We show a theoretical runtime analysis of the presented algorithm.
\item {\bf Experimental Analysis:} We show a parallel implementation of our algorithm using OpenMP and C++ which is over \(700\times\) faster than the naive parallel python code that internally uses optimized parallel math kernel libraries. The parallel implementation achieves \(67\times\) speedup on \(96\) cores sharing resources across \(4\) NUMA sockets of an Intel Cascade Lake machine w.r.t. its sequential run. 
\end{itemize}

\section{Background - Prior Work}
In this section, we discuss the necessary background and some prior research that are foundational for our work. The WMD attempts to capture this semantic similarity between documents. With words represented as vectors, each text document can be considered as a weighted point cloud of embedded words. Then, the distance between two text documents, say, \(A\) and \(B\) would be the minimum cumulative distance that words from document \(A\) need to travel to match exactly the point cloud of document \(B\). Distance between two words \(i\), \(j\) can be measured by the euclidean distance, \(m(i, j)\) = \({\parallel x_i - x_j \parallel}^2\) = sqrt(sum(pow(embedding[i] - embedding[j], 2))) between the high-dimensional vectors \(x_i\) and \(x_j\) associated with each word's word2vec representation. Essentially, \(m(i, j)\) gives the moving cost of word \(i\) to word \(j\) and a small distance indicates that words are closely related.   
 
Suppose, we have two documents \(A\) and \(B\) where A = ``Obama speaks to the media in Illinois'', and B = ``The President greets the press in Chicago''. These two sentences have intuitively the same meaning and therefore, these sentences should be ``more similar" to each other than, for example, to the sentence ``Amy Adams was in deepFake''. After throwing away the information about word order, capitalization and removing the frequent and uninformative stop-words (e.g., in, to, the), we get the following ``bag-of-words'' representation of \(A\) and \(B\): A = [`illinois', `media', `speaks', `obama'] and B = [`chicago', `greets', `president', `press']. Since \(A\) and \(B\) don't contain any of the same words, one can not look at the set intersection to measure sentence similarity. However, if the words are represented as vectors in the word-embedding space, it is expected that the word `obama' would be close to `president' and `Chicago' will be close to `Illinois' and \(m(media, press) < m(media, obama)\), where,
\(m(media, press) = sqrt(sum(pow(embeddings[media] - 
embeddings[press], 2)))\) \\
\(m(media, obama) = sqrt(sum(pow(embeddings[media] -
embeddings[obama], 2)))\).

\begin{wrapfigure}{r}{0.5\textwidth}
  \centering
  \includegraphics[width=0.5\textwidth]{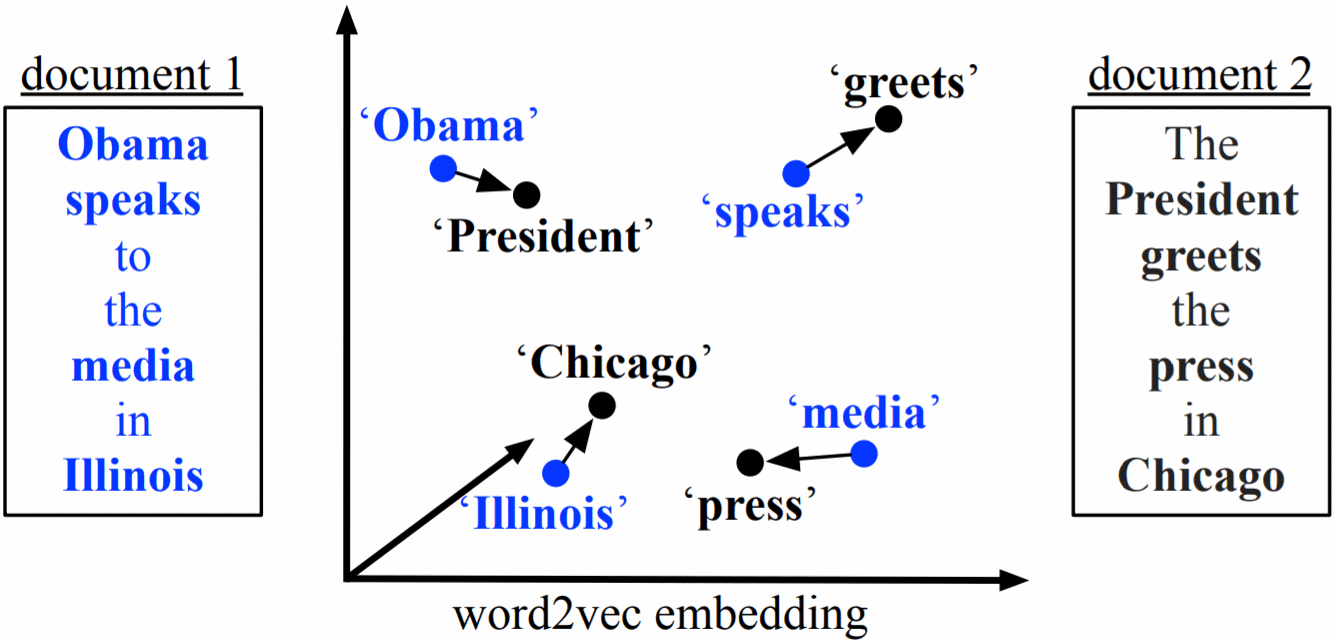}
  \caption{Example of a WMD calculation between two single-sentence documents \cite{WMD}.}
  \label{fig:WMD-principle}
\end{wrapfigure}  

Even though there is no overlap in words between the two documents, the word embedding provides the semantic association between disjoint words. Figure \ref{fig:WMD-principle} shows the WMD between documents \(A\) and \(B\) as the minimum cumulative distance that all non-filler words in the document \(1\) (blue) need to travel to exactly match the words in document \(2\) (black). Here, the WMD is the sum of the lengths of all black arrows. 

The WMD is similar to EMD and OTD that measure the cost of the optimal way to transport dirt from a set of source piles to a set of destination piles. EMD represents a family of well-studied problems in operations research for which several specialized solvers\cite{pele2009fast} have been developed. If the word embedding vectors of size \(w\) are considered as points in a \(w\)-dimensional space, then the distance between words \(i\) and \(j\) can be interpreted as the `cost' of transporting a unit of `mass' from point \(embedding[i]\) to point \(embedding[j]\). Then, sentences \(A\) and \(B\) can be considered as sets of points in that space, with a unit of mass piled on each point in \(A\) and \(B\). With this setting, the Word Mover's Distance would be the minimum cumulative cost of transporting all of the mass from points [\(embedding[i]\) for \(i\) in \(A\)] to points [\(embeddings[j]\) for \(j\) in \(B\)]. Note that mass can and might flow from a single point in \(A\) to multiple points in \(B\) and vise-versa because words can, of course, occur multiple times and can be mapped to different related words. The \(num\_words(A)\) \(\times\) \(num\_words(B)\) matrix that specifies the flow of mass from \(A\) to \(B\) is called the `transportation plan'.

If we have a database of \(5\)M documents, running a single query to compute the WMD of document \(A\) to all other \(5\)M documents would involve solving \(5\)M optimization problems of size \(O(V^3log(V))\) where \(V\) is the number of unique words (e.g., 100K) in the documents, which is quite costly. The original paper by Kushner et. al. \cite{WMD} used a flow-based approach to balance the total `in' and `out' flows of masses to compute the WMD. Many recent works follow a flow based technique to compute WMD \cite{10.1007/978-3-030-14799-0_11}, \cite{brokos2016using}, \cite{zhang2016building}, \cite{wu2018word}, \cite{tashu2018pair} which may take \(O(V^3log(V))\) time in the worst case. Several pruning ideas have been proposed in \cite{WMD} to speed up the document retrieval process that reduces the number of expensive WMD evaluations per query. However, the individual WMD computation is still prohibitively expensive in practice. 

Cuturi et. al. \cite{Cuturi:2013:SDL:2999792.2999868} proposed an approximation to the Optimal Transportation Distance that reduced the cost per query for the Earth Movers Distance to \(O(V^2)\) by adding an entropy regularization term: basically, the optimization problem is regularized with this entropic term following the maximum-entropy principle. This encourages the solution to lie `close' to the (trivial) transportation plan that sends equal mass from each point in \(A\) to each point in \(B\). `Closeness' is measured by Kullback-Leibler (KL)-divergence - a measure of how one probability distribution (e.g., word frequency of one document) is different from a second reference probability distribution. Using this entropic term, the optimization problem is converted from a general linear programming (LP) problem into a convex problem, which can be solved using the Sinkhorn-Knopp matrix scaling algorithm \cite{Knight:2008:SAC:1404637.1404647}. The authors in \cite{Cuturi:2013:SDL:2999792.2999868} called the new approximated distance as the Sinkhorn distance and proved that for large enough entropy, the Sinkhorn distance is equivalent to the optimal transportation distance (hence, the proof is omited here). Furthermore, the Sinkhorn distance is symmetric and satisfies triangle inequality and is a metric. The added penalty/entropy makes the problem substantially easier to solve and one can use the fast Sinkhorn-Knopp matrix scaling algorithm which is also easily parallelizable. This formulation makes the problem a non-flow problem.

\section{Mapping of OTP Algorithm to WMD Algorithm}
Algorithm \ref{Algo1} shows how the Sinkhorn distance can be computed for the Optimal Transportation Problem \cite{Cuturi:2013:SDL:2999792.2999868}. In the algorithm, \({{V^\lambda}_M}(r,c)\) represents the approximate optimal transportation distance or the Sinkhorn distance between two probability distributions ( \(r\) for the query and \(c\) for target). In case of ``one to many" distance computation, \(c\) would hold the distribution of many target documents. In the context of WMD, \({{V^\lambda}_M}(r,c)\) would be equivalent to the WMD between two documents, \(r\) would be the word frequency vector/histogram of the input/source document and \(c\) would be the word frequency vector/histogram of the target document. If we have multiple target documents, \(c\) would then represent an array of vectors (i.e., a matrix) of word frequencies of all target documents with each column of \(c\),\(c[:j]\) denoting the word frequency vector of target document \(j\). The \(\lambda\) would denote the regularizing entropy parameter, \(M\) would denote the pair-wise transport cost matrix, e.g, the euclidean distance matrix among each pair of words in the dictionary. 

\vspace{-2pt}
\begin{algorithm}[h]
\SetAlgoLined
\KwData{\(M, \lambda, r, c.\)} \KwResult{\({d_M}^\lambda (r, c)\) }
  \(I=(r>0); r=r(I); M=M(I,:); K=exp(-\lambda*M)\) 
  Set \(x=ones(length(r),size(c,2))/length(r)\)    
  \While{x changes}
  {
  \(x=diag(1./r)*K*(c.*(1./(k^T*(1./x))))\)  
  }
  \(u=1./x; v=c.*(1./(k^T*u))\)  
  \\
  \({d_M}^\lambda (r, c)=sum(u.*((K.*M)*v))\) 
 \caption{Computation of \({d_M}^\lambda (r, c)\)  using Sinkhorn-Knopp\cite{Cuturi:2013:SDL:2999792.2999868}.}
 \label{Algo1}
\end{algorithm}
 \vspace{-4pt}
 
When \(\lambda > 0\), the solution \(P^\lambda\) is unique by strict convexity of the minus entropy and is of the form \(u_i e^{-\lambda. m_{i,j}} v_j\), \(u\) and \(v\) are two non-negative vectors uniquely defined up to a multiplicative factor.
\begin{equation}
\exists u, v > 0_d : P^\lambda = diag(u) e^{-\lambda M} diag(v).
\end{equation}

Given \(e^{-\lambda} M\) and marginals \(r\) and \(c\), the original paper \cite{Cuturi:2013:SDL:2999792.2999868} showed that it is sufficient to run enough iterations of Sinkhorn Knopp's matrix scaling algorithm to converge to a solution \(P^\lambda\) of that problem. With one source document and \(N\) target documents, it takes \(O(V^2N t)\) linear algebra operations to compute the WMD to all documents where \(t\) is the number of iterations to converge.

{\bf Motivations:} WMD provides lower error rate compared to other standard metrics such as bag-of-words (BOW), term frequency-inverse document frequency (TFIDF), Latent Semantic Indexing (LSI), Latent Dirichlet Allocation (LDA) and so on for document classification and very recently, WMD has been used in various types of text processing applications \cite{10.1007/978-3-030-14799-0_11}, \cite{brokos2016using}, \cite{zhang2016building}, \cite{wu2018word}, \cite{tashu2018pair}. The above way of computing WMD provides an asymptotic performance boost over the original flow-based approach \cite{WMD}. For example: for the k-nearest neighbor algorithm, the computation cost would reduce from \(O(V^3 \log V N t + k\log N)\) to \(O(V^2N t + k \log N)\). A very recent paper \cite{Li:2019:CES:3308558.3313397} used this approach as part of their work on short-text clustering. However, they used a sequential approach to compute WMD and their algorithmic description appeared to be complicated to apprehend. 

{\bf Contributions:} In this paper, we adapt to the Optimal Transportation Algorithm \cite{Cuturi:2013:SDL:2999792.2999868} to compute WMD in a way that is simple, easy to follow and implement. Despite the cost reduction from \(O(V^3 \log V N t)\), to \(O(V^2N t)\), the quadratic \(O(V^2N t)\) computational cost is still expensive, especially if the number of unique vocabulary words is in the order of millions or billions. Therefore, parallelizing the algorithm to compute WMD would be an obvious choice to make the computation faster and also more practical. With that goal in mind, in this paper, we present a high-performance shared memory parallel algorithm to compute WMD. 

\section{Parallel Sinkhorn Word Movers Distance Algorithm}
In this section, we present our parallel Sinkhorn-Knopp Algorithm to compute WMD efficiently. We assume that we have one source document and many target documents. We start with a python implementation of Algorithm \ref{Algo1} with relavant modifications for WMD. Then we present an optimized parallel version of that algorithm that we implemented in C/OpenMP. 

\begin{wrapfigure}{r}{0.5\textwidth}
  \centering
  \includegraphics[width=0.5\textwidth]{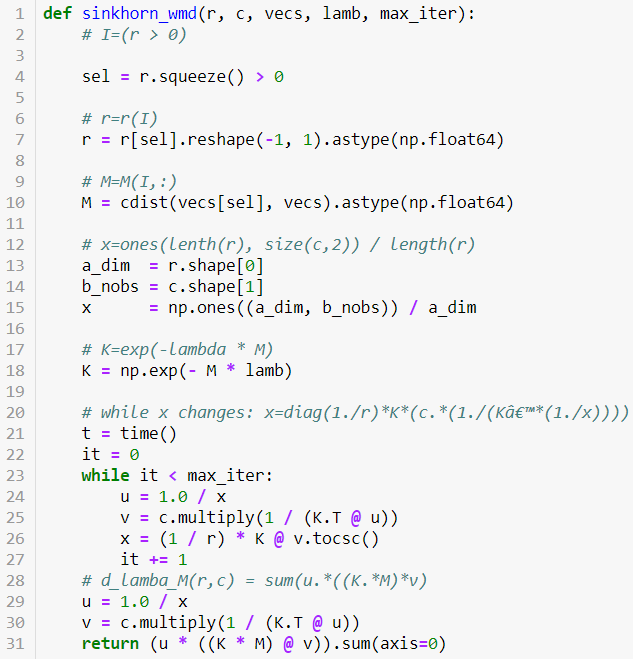}
  \caption{Python Implementation of Algorithm \ref{Algo1}.}
  \label{fig:WMD-python}
\end{wrapfigure}

\paragraph{A Python Implementation}
Figure \ref{fig:WMD-python} shows a python implementation of Algorithm \ref{Algo1}. The sinkhorn\_wmd function takes \(r\), \(c\), \(vecs\), \(Lamda$ and \(max\_iter$ as inputs. Originally, \(r\) is a sparse vector of size |vocabulary| representing a histogram of word frequencies in the query/source document. More specifically, \(r[i]\) denotes the normalized count of \(i^{th}\) vocabulary word in the source document. Entries in \(r\) are normalized so that \(sum(r) = 1$. The \(c\) is a sparse \(vocabulary\_size \times num\_docs\) size matrix where \(c[i,j]\) denotes the normalized frequency/count of the \(i^{th}\) word in the \(j^{th}\) target document. The columns of \(c\) are normalized so that sum of normalized word frequency in a particular target document produces \(1\), i.e., \(c[:,j] = 1\). The \(vecs\) is a \(vocabulary\_size \times word\_embedding\_size\) dense matrix where the \(i^{th}\) row of \(vecs\) provides the word2Vec (or BERT\cite{devlin2018bert} or ElMo\cite{peters2018deep}) word-embedding vector of \(i^{th}\) word in the dictionary/vocabulary set. The \(lamda\) is the regularization parameter and \(max\_iter\) is the maximum number of iterations to find the solution.

In python, the \(*\) operator works as an element-wise multiplication and \(@\) operator performs the canonical matrix multiplication operation. The python function sinkhorn\_wmd starts by selecting the nonzero entries of \(r\), i.e., selecting only those words from the vocabulary that appears in the source document. After that, on line 6, it overwrites \(r\) with one that contains only the non-zero entries. Let, \(v\_r\) denotes the number of non-zero entries in \(r\). Next, it computes the euclidean distance matrix, \(M\) such that \(m(i, j)\) = \(sqrt(sum(pow(embedding[i] - embedding[j], 2)))\), where \(i\) denotes a word that appears in the source document and \(j\) denotes a word that appears in the dictionary. Note that, we lazily compute \(M\) based on the nonzero entries of the source document \(r\), instead of computing the full \(vocabulary\_size \times vocabulary\_size\) distance matrix. On line 14, we create the \(x\) matrix of size \(v\_r \times num\_docs\) assuming an equal amount of mass/weight in all points/words. Next, it computes other matrices that need to be computed only once: \(K[i,j]=exp^{Lamda . M[i,j]}\), \(K\_over\_r = (1 / r) * K\), \(KT = K.T\) and \(KM = (K * M)\). Lamda is passed to the function after being negated. Next, it runs the solver loop to iteratively reach close to the optimal solution (i.e., WMV). It computes \(u = 1/x\) (i.e., \(u[i,j]=1/x[i,j]$), then, \(v = c.multiply(1 / (KT @ u))\) and then, \(x = K\_over\_r @  v.tocsc()\). Here, \(c.multiply\) is an element-wise multiplication and \(v.tocsc\) transposes the matrix in sparse format. After the max\_iter number of iterations, the while loop exits. In an ideal scenario, one would want to iterate as long as there is any change in \(x\) during an iteration of the loop. After exiting the while loop, it computes \(u = 1.0 / x\), \(v = c.multiply(1 / (KT @ u))\), followed by \(WMD=(u * ((KM) @ v)).sum(axis=0)\) operation. Here, \(WMD\) is a vector of size \(num\_docs\), where, WMD[i]=Sinkhorn\_distance(src\_doc, target\_doc[i]).
\paragraph{Dataset}
In this paper, we use a precalculated word embeddings as an input that is trained by Google on a very large number of documents scraped from the internet and captures a lot of information about the meanings of words. The dictionary size (e.g., vocabulary\_size) is \(100,000\) words and the word-embedding vector size is \(300\). Therefore, with fp64 numbers, the dictionary size is \(100,000 \times 300 \time 8$=$0.24GB\). This precalculated word-embeddings is a subset of the \verb|crawl-300d-2M.vec|\footnote{See for instance \url{https://www.kaggle.com/yekenot/fasttext-crawl-300d-2m\#crawl-300d-2M.vec}} word embedding. We use documents from \verb|dbpedia.train.gz| database\footnote{See for instance \url{https://www.kaggle.com/lotuswhl/dbpediafromfasttext}} as our source and target documents. We use the first \(5000\) documents as our target documents and first \(10\) documents as source/query documents. The properties of the matrices are as follows:
\begin{itemize}
\item \texttt{c}: a sparse matrix with \(100,000\) \(\times\) \(5000\) fp64 elements, holding normalized word frequency in the documents. Approx. \(0.035\) \% of the total entries are non-zero.
\item \texttt {vecs}: a dense matrix with \(100,000\) \(\times\) \(300\) fp64 elements, holding the word embeddings.
\item \texttt{r}: a sparse vector with \(100,000\) elements, holding the word frequency of the input document.
\end{itemize}

\begin{table*}[htb!]
\centering
\vspace{-10pt}
  \caption{Profile of the python code on an Intel\textregistered{} Xeon\textregistered{} system.}
  \scalebox{0.8}
  {    
    \begin{tabular} {|r|l|r|}
      \hline
      \multicolumn{1}{|c}{\bf Runtime} \% & \multicolumn{1}{|c}{\bf Code line} & \multicolumn{1}{|c|}{\bf Potential Kernel} \\ \hline
      0 \% & \texttt{sel = r.squeeze() > 0}  & \\ \hline
      0 \%& \texttt{r = r[sel].reshape(-1, 1).astype(np.float64)}  & \\ \hline
      1.4 \%& \texttt{M = cdist(vecs[sel], vecs).astype(np.float64)}  & \\ \hline
      0 \%& \texttt{a\_dim  = r.shape[0]}  & \\ \hline
      0 \%& \texttt{b\_nobs = c.shape[1]}  & \\ \hline
      0 \%& \texttt{x = np.ones((a\_dim, b\_nobs)) / a\_dim}  & \\ \hline
      0 \%& \texttt{K = np.exp(- M * lamb)}  & \\ \hline
      0 \%& \texttt{p=(1 / r) * K}  & \\ \hline
      0 \%& \texttt{it = 0}  & \\ \hline
      0 \%& \texttt{while it < max\_iter}  & \\ \hline
      0 \%& \texttt{u = 1.0 / x}  & \\ \hline
      0 \%& \texttt{KT=K.T}  & \\ \hline
      91.9 \%& \texttt{v = c.multiply(1 / (KT @ u))}  & Sparse$\times$Dense$\times$Dense matrix\\ \hline
      0.1 \%& \texttt{v\_csc=v.tocsc()}  & \\ \hline
      0.5 \%& \texttt{x = K\_over\_r v\_csc}  & Dense$\times$Sparse matrix\\ \hline
      0 \%& \texttt{it += 1}  & \\ \hline
      0 \%& \texttt{u = 1.0 / x}  & \\ \hline
      6.1 \%& \texttt{v = c.multiply(1 / (K.T @ u))}  &  Sparse$\times$Dense$\times$Dense matrix\\ \hline
      0 \%& \texttt{return (u * ((K * M) @ v)).sum(axis=0)}  & Dense$\times$Sparse matrix\\ \hline
  \end{tabular} 
  }
  \label{table:sinkhorn_profile}
  \vspace{-10pt}
\end{table*}

\paragraph{Profiling on CPU}
We ran the Python code shown in Figure \ref{fig:WMD-python} on a 2-socket ``Cascade Lake'' generation Intel\textregistered{} Xeon\textregistered{} Platinum 8280 system with 28 cores per socket clocked at \(2.70\) GHz. Although the python code appears to be sequential, it internally used \(48\) threads (observed using \emph{top} command on linux terminal) and took approximately \(64\) seconds for a source document with \(19\) words (i.e., \(19\) non-zero entries in \(r$). Those threads were used by Math Kernel Library (MKL)/Blas library calls. Notice that, the python implementation requires dense matrix multiplication (@) on large arrays \((K.T @ u)\) of size \((100,000 \times 19)\) @ \((19 \times 5000)\) and then an element-wise multiplication ($*$) by a sparse matrix \(c\) with the resultant dense matrix, \(c.multiply(K.T @ u)\) whereas the sparse matrix \(c\) has only \(173087\) non-zero values out of \((100,000 \times 5000)\) possible entries (i.e., density = \(0.00346174\) \%). The initial Python profiling of the workload using \verb|cProfile| and \verb|line profile| tools is shown in table \ref{table:sinkhorn_profile}. The key computational kernel is the (Sparse * Dense @ Dense) matrix multiplication which takes upto \(98\%\) of the total time. The (Dense @ Sparse) matrix multiplication is insignificant in terms of runtime as shown in Table \ref{table:sinkhorn_profile}. We further profiled the code using Intel's Vtune profiler that highlights that the MKL library calls take the most amount of time in the code.

\begin{figure}[hptb!]
\vspace{-10pt}
  \centering
  \includegraphics[width=0.8\textwidth]{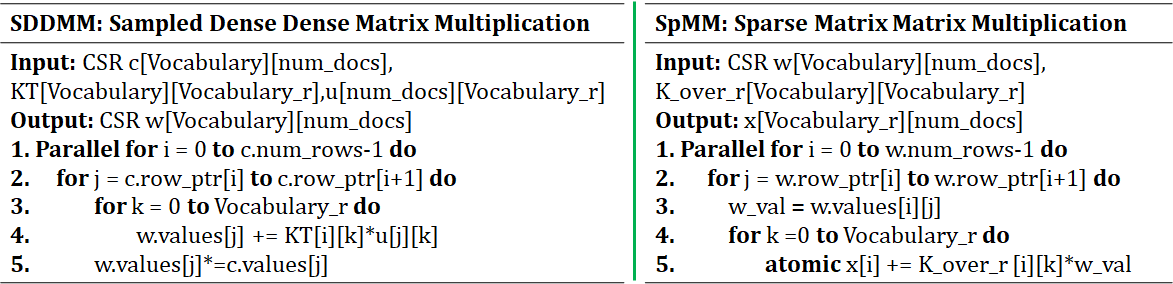}
  \caption{SDDMM and SpMM Kernels.}
  \label{fig:SDDMM}
  \vspace{-10pt}
\end{figure}

\paragraph{Dense to Sparse Kernel}
In the Python implementation, the \(c.multiply(1 / (KT @ u)\) operation, i.e., the dense matrix-multiplication followed by the sparse element-wise multiplication is not only highly costly but also performs several unnecessary multiplications and additions that could be avoided easily. We converted the entire (sparse * dense @ dense) to a sparse selection of dense matrix multiplication kernel (e.g., a SDDMM kernel shown in Fig. \ref{fig:SDDMM} left), that performs a dot product only for that row ($i$) and a column ($j$) for which \(c[i,j]\) is non-zero. That way, instead of doing dense matrix multiplications and then filtering out most of them by a sparse matrix, we first select which dot products are needed using the non-zero \(c[i,j]$s and then do those dense dot products. The resultant matrix of the SDDMM operation is a sparse matrix which is of the same size as \(c\). Next, we do a dense and sparse matmul \((x = K\_over\_r @ v\_csc)\) using an SpMM (Fig. \ref{fig:SDDMM} right) kernel. 

\begin{figure*}[htb!]
  \centering
  \includegraphics[width=0.485\textwidth]{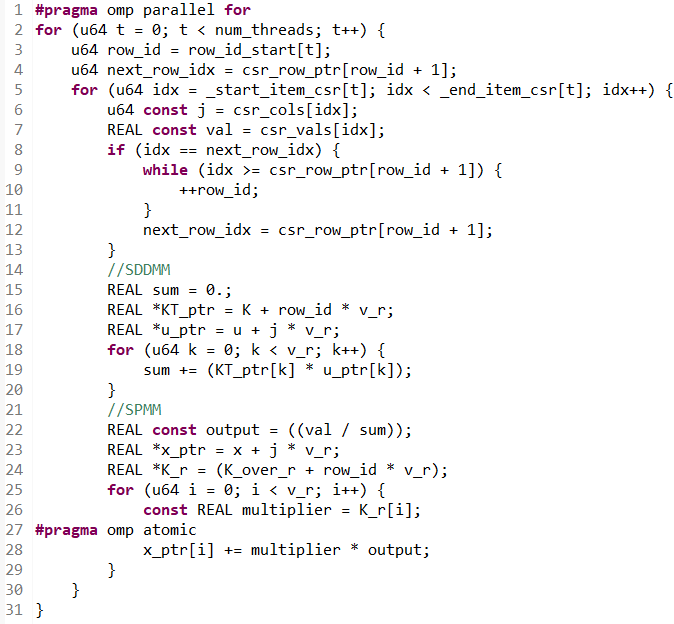}
  \includegraphics[width=0.485\textwidth]{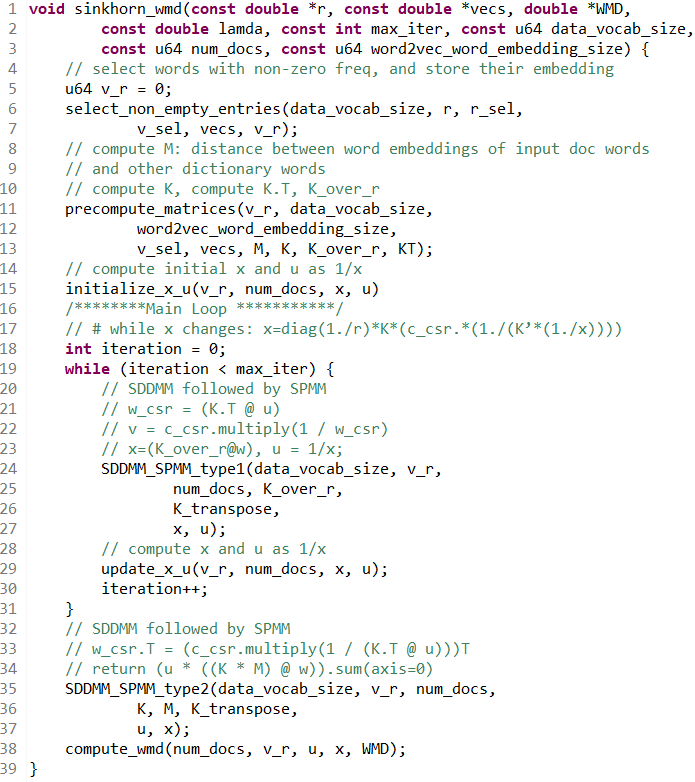}
  \caption{Left: Code snippet of SDDMM\_SpMM kernel, Right: Snippet of the C Implementation.}
  \label{fig:SDDMMandSpMM}
\end{figure*}

\paragraph{The SDDMM\_SpMM Kernel} We fused the SDDMM and SpMM kernels to create a new sparse matrix kernel named SDDMM\_SpMM. The benefits of SDDMM\_SpMM are \(1)\) it frees us from iterating twice over the CSR rows and column ids, \(2)\) the output values from SDDMM can be fed directly to the SpMM and would not need to be stored in memory to be used in later SpMM. Additionally, data could be transposed on the fly to ensure unit-stride data accesses. Notice that the \(K\_over\_r\), \(K.T\), \(M\) matrices can be pre-computed once and reused over and over again during the while loop iterations. Figure \ref{fig:SDDMMandSpMM} shows a code snippet of SDDMM\_SpMM, that fuses the standard SDDMM and SpMM kernels. For {\bf load-balancing}, we have divided the number of non-zeros in \(c\) matrix evenly among the threads and each thread in parallel determines its starting exploration point inside the CSR using a binary search which guarantees an equal work distribution across threads.

\paragraph{C Implementation}
Based on the above observations, we implemented a \(C\) version of the python code where we applied the following optimizations: 
\begin{itemize}
\item Elimination of all dense matmul.
\item Fusion of SDDMM and SpMM Kernel to SDDMM\_SpMM kernel.
\item On the fly transpose for unit stride data access.
\item Optimized pointer arithmetics to avoid overheads of address calculation.
\item Basic unrolling, other standard optimizations, vectorizations, parallelization.
\item Function inlining, common sub-expression eliminations.
\end{itemize}

Figure \ref{fig:SDDMMandSpMM} (right) shows a skeleton of implementation of the Sinkhorn-Knopp algorithm. All sub-functions were parallelized. In SDDMM\_SpMM (left), we divide the number of non-zeros in the \(c\) matrix equally among all workers and we compute each non-zero element of \(c.multiply(1 / (KT @ u)\) in parallel but mutually exclusively and hence we do not need any atomics there. 
\vspace{-5pt}\subsection{Theoretical Analysis}
If the word-embedding vector for each dictionary word is of size \(w\), number of words in the source doc is \(V\_r\), total number of words in the dictionary is \(V\), number of target documents is \(N\), and the number of total non-zeros in the CSR is \(nnz\), then the select\_non\_zero\_entries function would take \(O(V + {V\_r . w\over p} + \log p)\) time where \(p\) is the number of threads. The precompute\_matrices would take \(O({V\over p} . V\_r . w + {V\over p} . V\_r + \log p)\) and Initialize\_x\_u \(O({{N \over p}.V\_r } + \log p)\) time. 

\begin{table}[htbp]
  \centering
  \caption{Asymptotic Runtime Cost}
    \begin{tabular}{l|l|l}
    \toprule
          & 1 to N WMD Cost & 1 to 1 WMD Cost \\
          \midrule
    Total cost  & \(O ({V . V\_r. w  \over p }+ t* {nnz. V\_r \over p}$) & \(O ({V\_x . V\_r. w  \over p} + t* {V\_x . V\_r \over p})\) \\
    \bottomrule
    \end{tabular}%
  \label{tab:cost}%
\end{table}%

To determine the starting point in CSR, each thread need to spend \(O(\log V)\) time. Inside the while loop, the SDDMM\_SPMM (both type1 and type2) takes \(O({V\over p} + {nnz\over p} . V\_r +\log p)\), and update\_x\_u takes \(O({N.{V\_r \over p}} + \log p)\). Finally, the compute\_WMD takes \(O(N. {V\_r \over p} + \log p)\). The \(\log p\) terms account for the thread spawning and barrier overheads. Table \ref{tab:cost} shows the asymptotic cost of running our proposed parallel algorithm for computing WMD of one document against N other target documents (1 to N). The table also shows the average cost per target document (1 to 1) where the target document has \(V\_x\) unique words and \(t\) is the number of iterations for convergence.  

\begin{wraptable}{r}{0.5\textwidth}
  \centering
  \caption{System Spcifications}
  \resizebox{0.5\textwidth}{!}
  {
    \begin{tabular}{|c|c|c|}
    \toprule
    \textbf{Platfroms} & \multicolumn{1}{p{7em}|}{\textbf{CLX0}} & \multicolumn{1}{p{7em}|}{\textbf{CLX1}} \\
    \midrule
    \textbf{Model} & \multicolumn{1}{p{7em}|}{\textbf{Intel(R) Xeon(R) Platinum 8280 CPU @ 2.70GHz}} & \multicolumn{1}{p{7em}|}{\textbf{Intel(R) Xeon(R) Platinum 9242 CPU @ 2.30GHz}} \\
    \midrule
    \textbf{Cpu MHz} & \multicolumn{1}{p{7em}|}{1800}  & \multicolumn{1}{p{7em}|}{1600} \\
    \midrule
    \textbf{L1d} & \multicolumn{1}{p{7em}|}{32KB} & \multicolumn{1}{p{7em}|}{32KB} \\
    \midrule
    \textbf{L2} & \multicolumn{1}{p{7em}|}{1024KB} & \multicolumn{1}{p{7em}|}{1024KB} \\
    \midrule
    \textbf{L3} & \multicolumn{1}{p{7em}|}{ 39.4MB } & \multicolumn{1}{p{7em}|}{36.6MB} \\
    \midrule
    \textbf{MemAvailable} & \multicolumn{1}{p{7em}|}{190GB} & \multicolumn{1}{p{7em}|}{390GB} \\
    \midrule
    \textbf{\#Cores per socket} & \multicolumn{1}{p{7em}|}{28} & \multicolumn{1}{p{7em}|}{24} \\
    \midrule
    \textbf{\#Numa sockets} & \multicolumn{1}{p{7em}|}{2} & \multicolumn{1}{p{7em}|}{4} \\
    \bottomrule
    \end{tabular}%
    }
  \label{tab:sys-spec}%
\end{wraptable}%
\section{Experimental Results}
We implemented our parallel Sinkhorn-WMD algorithm in C/C++/OpenMP and compiled the program using the Intel icc version 19.0.2.187 (gcc version 4.8.5 compatibility) with the following compiler flags: \texttt{\scriptsize -O3 -fopenmp  -xHost -g -restrict -std=c++11 -finline -unroll -ansi-alias -qopt-subscript-in-range}. We used two state-of-the-art Intel Xeon systems codenamed Cascade Lake (CLX0 and CLX1). Table \ref{tab:sys-spec} shows the machine specifications. The operating system was CentOS Linux 7 (Core). It was not obvious from specs alone whether one would be faster than the other, and the expectation was that they should perform equally well. First, we used the same \(19$-word document as the input source that we used to profile the original python code. For this input, the python code takes around \(64\) sec on the CLX0 system, and surprisingly, it takes only \(0.091\) second {\textbf($700\times\) faster!)} on a single socket of the CLX0 machine. 


\begin{figure*}[htb!]
  \centering
  \begin{tabular}{|c|c|c|}
  \includegraphics[width=0.32\textwidth]{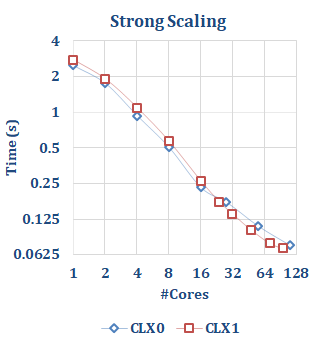}&
   \includegraphics[width=0.32\textwidth]{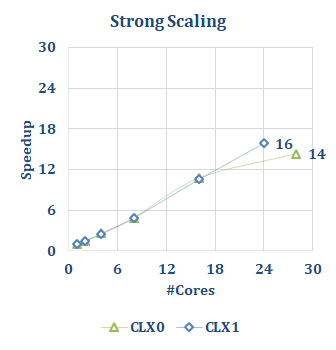}&
\includegraphics[width=0.32\textwidth]{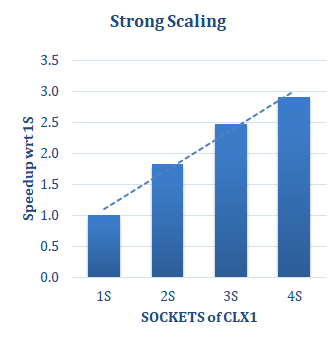}\\
  \end{tabular}
    \caption{1) Run time across sockets of CLX1 and CLX0. 2)Strong Scaling in One Socket of CLX1 and CLX0. 3)Strong Scaling in across Sockets of CLX1.}
  \label{fig:multisockettime}
\end{figure*}

\begin{figure*}[t!]
  \centering
  \begin{tabular}{|c|c|}
\includegraphics[width=0.4\textwidth]{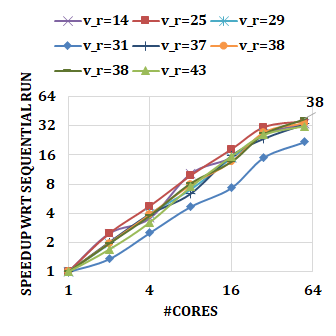}&
  \includegraphics[width=0.4\textwidth]{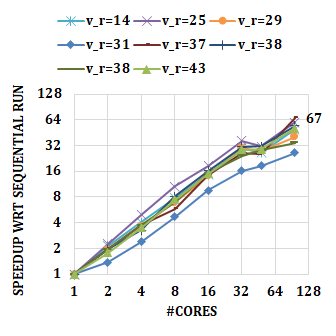} 
 \end{tabular}
 \caption{Strong Scaling on multiple input files on CLX0 and CLX1.}
  \label{fig:multiTimeCLX}
\end{figure*}

Next, we took a \(43$-word source document as input and computed WMD with all \(5000\) target documents containing \(100,000\) words. Figure \ref{fig:multisockettime} shows the strong scaling of the parallel Sinkhorn-WMD algorithm. The figure shows that the algorithm scaled nicely across sockets on both CLX0 and CLX1 machines. Within a socket, it obtained \(14\times\) speedup on \(28\) cores on CLX0 and for CLX1, the speedup was \(16\times\) on \(24\) cores. On CLX1, the code achieved \(3\times\) speedup on \(4\) sockets compared to its 1 socket run. The Intra-socket speedup of CLX1 appears to be better than that of CLX0 most likely due to having larger memory. 

Figure \ref{fig:multiTimeCLX} shows the runtime scaling of our parallel Sinkhorn-WMD algorithm on multiple source documents ran at once. In the Figure, \(v\_r\) stands for the number of words in the source document. Figure \ref{fig:multiTimeCLX} shows that for CLX0, the maximum speedup obtained was for \(v\_r=38\) and it was about \(38\times\) speedup on two-sockets ($56$-cores) of CLX0. The \(v\_r=31\) has the worst speedup among all because it was the very first source/query file in the input list and had affected by the cold misses. The maximum speedup obtained on CLX1 was for \(v\_r=37\) and it was about \(67\times\) speedup on four-sockets ($96$-cores) of CLX1. We also see a clear dip after crossing two-sockets ($48$-cores). Similar to the prior case, the \(v\_r=31\) had the worst speedup among all because it got affected by the cold misses.

\section{Computation of Euclidean Distance}
When we analyzed our algorithm using Intel Vtune, we found that most of the time is spent in kmp\_wait (libiomp) and other libc functions, mainly outside of the region of interest (Sinkhorn\_WMD kernel). The most time-consuming user function was the Euclidean distance function that resembles a dot product, among vectors of size \(w\times w\), where \(w=300\) for our case. One way to potentially improve the performance of Euclidean distance is to re-structure the code in a way so that it becomes like a matrix-multiplication-like kernel. The Euclidean distance computation can be transformed to appear like a dense matrix multiplication (i.e., \(C=A@B$) where, A=v\_r \(\times\) w=|$19\times300$|, B=w \(\times\) vecs.T=|$300\times100,000$| and M=euclidean\_distance(v\_r, vecs)=|$19\times100,000$|. Unlike a standard matmul where each update requires 2 floating-point operations (FLOPS), this kernel would require \(3\). Then, a highly optimized cache-efficient matrix-multiplication kernel can be modified appropriately to implement that function and by that, the bandwidth requirement can be reduced. 
\begin{wrapfigure}{r}{0.5\textwidth}
  \centering
  \includegraphics[width=0.4\textwidth]{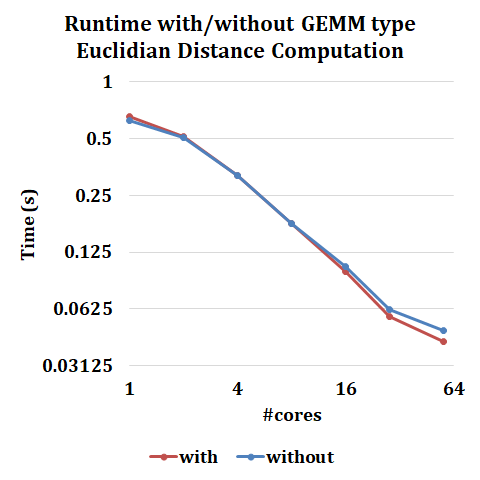}
  \caption{Runtime with/without GEMM type Euclidean Distance Computation.}
  \label{fig:EDCLX0}
\end{wrapfigure}

Figure \ref{fig:EDCLX0} shows the performance difference of two implementations one of which uses a blocked matrix-multiplication type Euclidean distance computation and another uses a dot-product type computation. We blocked the \(j\) loop that iterates over the \(100,000\) vocabulary words and the \(i\) loop that iterates over the vocabulary of the query document. We did not block the \(k\) loop that iterates over the word-embedding vector. The top \(j\) loop was parallelized. With such an implementation, for the \(19$-word document, we see almost no difference in runtime between the two versions till 8 cores and after that, we see a slight improvement. It might be due to the fact that the query document is a tall and skinny matrix (19 \(\times\) 300) which limits the efficiency. Furthermore, we use the modified matrix-multiplication-like kernel to not only compute matrix \(M\) but also \(K\) and \(K\_over\_r\) matrices at once which increases the working set-size on cache compared to a typical matrix multiplication that handles only three matrices.


\section{Conclusion}
In this paper, we have presented an efficient parallel sparse algorithm to compute the Sinkhorn Word Movers Distance (WMD) that replaces the original dense compute kernels with sparse kernels. We introduce a new kernel called SDDMM\_SpMM that fuses the Sparse selection of Dense-Dense matrices and Sparse matrix Dense matrix multiplication kernels. WMD has been shown to get better accuracy for many state-of-the-art Natural Language Processing (NLP) applications to understand the semantic similarity between two documents or even images. Our results show that although the current trend is to use python programming language for machine learning, it can be \(700\times\) to \(800\times\) slower than a simple C/OpenMP implementation. We were able to get as good as \(67\times\) speedup on \(96\) cores on state-of-the-art Intel multicore machine sharing resources across NUMA domains. The research presented in this paper has practical significance, because, 1) it presents a new algorithm to compute Word Movers Distance that has better asymptotic theoretical bounds and is also practically efficient, 2) the parallel version is scalable and runs two orders of magnitude faster compared to a python implementation on state-of-the-art CPUs and also simple enough to implement for data-scientist. We would love to share the code upon request.

\section{Acknowledgement}
The work was done as part of DARPA HIVE and SDH project. We would like to thank Mathieu Gontier, Intel, Belgium, for his help during the porting of Sinkhorn to a special purpose accelerator which is outside the scope of the work presented in this paper. We plan to publish that result when suitable.
\newpage
\bibliographystyle{plain}
\bibliography{neurips_2020}

\section*{Appendix}
A {\bf novelty} of our algorithm is that whereas the original Sinkhorn algorithm is dense compute-heavy, our proposed algorithm is sparse compute-heavy which gave us performance boost even in the sequential version. Most of the recent research has used a python- or Matlab-based implementation to compute WMD. These programming languages are popular due to their simplicity and availability of commonly used libraries. These programming languages give productivity but often sacrifices performance in terms of runtime compared to native C/C++ languages. Furthermore, the use of pre-defined libraries restrains from performing fusion of kernels and other custom optimizations. In this paper, we present a simple parallel algorithm that can be easily implemented in C/C++ to compute WMD. Implementation of our parallel algorithm is orders of magnitude (\(700\times\) to \(800\times\)) faster than the python implementation of Algorithm \ref{Algo1} that internally uses optimized parallel math kernel library.
\section{Other Improvement Ideas}
In this paper, we have not applied standard tiling optimizations in the SDDMM or SpMM kernels. If we assume that all matrices can be loaded from cache, the runtime of our new SDDMM\_SpMM kernel can be improved further. There are efficient approaches proposed in the literature to tile SDDMM and SpMM such as the efficient adaptive sparse tiling \cite{hong2019adaptive} which can be adapted to make those kernels cache-efficient. However, the speedup might be limited if most of the matrices are tall and skinny. Any kind of specialized and smart word-embedding can be used to compute the WMD. Recently in NLP, positional and contextual embeddings are being generated to capture the semantic meaning of the text with greater accuracy. Those embeddings can be used to compute WMD as well.
\end{document}